\documentclass[10pt,twocolumn,letterpaper]{article}

\usepackage{cvpr}
\usepackage{times}
\usepackage{epsfig}
\usepackage{graphicx}
\usepackage{amsmath}
\usepackage{amssymb}

\usepackage{pdfpages}

\usepackage[ruled, linesnumbered]{algorithm2e}

\usepackage[pagebackref=true,breaklinks=true,letterpaper=true,colorlinks,bookmarks=false]{hyperref}

\renewcommand{\a}{\mathcal{A}}

\newtheorem{remark}{Remark}

\cvprfinalcopy 

\def\cvprPaperID{3708} 

\begin{document}

\title{Exploiting Symmetry and/or Manhattan Properties for \\ 3D Object Structure Estimation from Single and Multiple Images}

\author{Yuan Gao\\
Tencent AI Lab, Shenzhen, China\\
{\tt\small ethanygao@tencent.com}
\and
Alan L. Yuille\\
Johns Hopkins University, Baltimore, MD\\
UCLA, Los Angeles, CA \\
{\tt\small alan.yuille@jhu.edu}
}

\maketitle

\begin{abstract}
Many man-made objects have intrinsic symmetries and Manhattan structure. By assuming an orthographic projection model, this paper addresses the estimation of 3D structures and camera projection using symmetry and/or Manhattan structure cues, which occur when the input is single- or multiple-image from the same category, \eg, multiple different cars. Specifically, analysis on the single image case implies that Manhattan alone is sufficient to recover the camera projection, and then the 3D structure can be reconstructed uniquely exploiting symmetry. However, Manhattan structure can be difficult to observe from a single image due to occlusion. To this end, we extend to the multiple-image case which can also exploit symmetry but does not require Manhattan axes. We propose a novel rigid structure from motion method, exploiting symmetry and using multiple images from the same category as input. Experimental results on the Pascal3D+ dataset show that our method significantly outperforms baseline methods.
\end{abstract}

\section{Introduction}

\begin{figure}[t]
\centering
\includegraphics[width=0.3\linewidth]{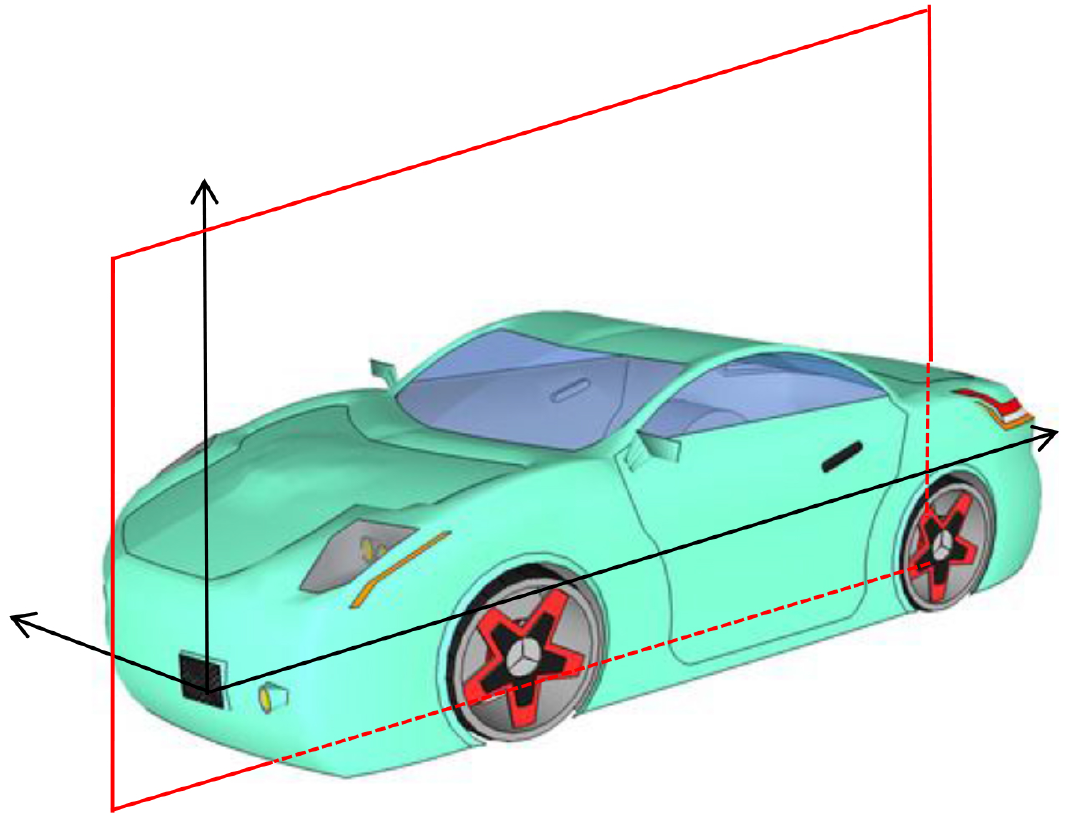}
\hspace{15mm}
\includegraphics[width=0.3\linewidth]{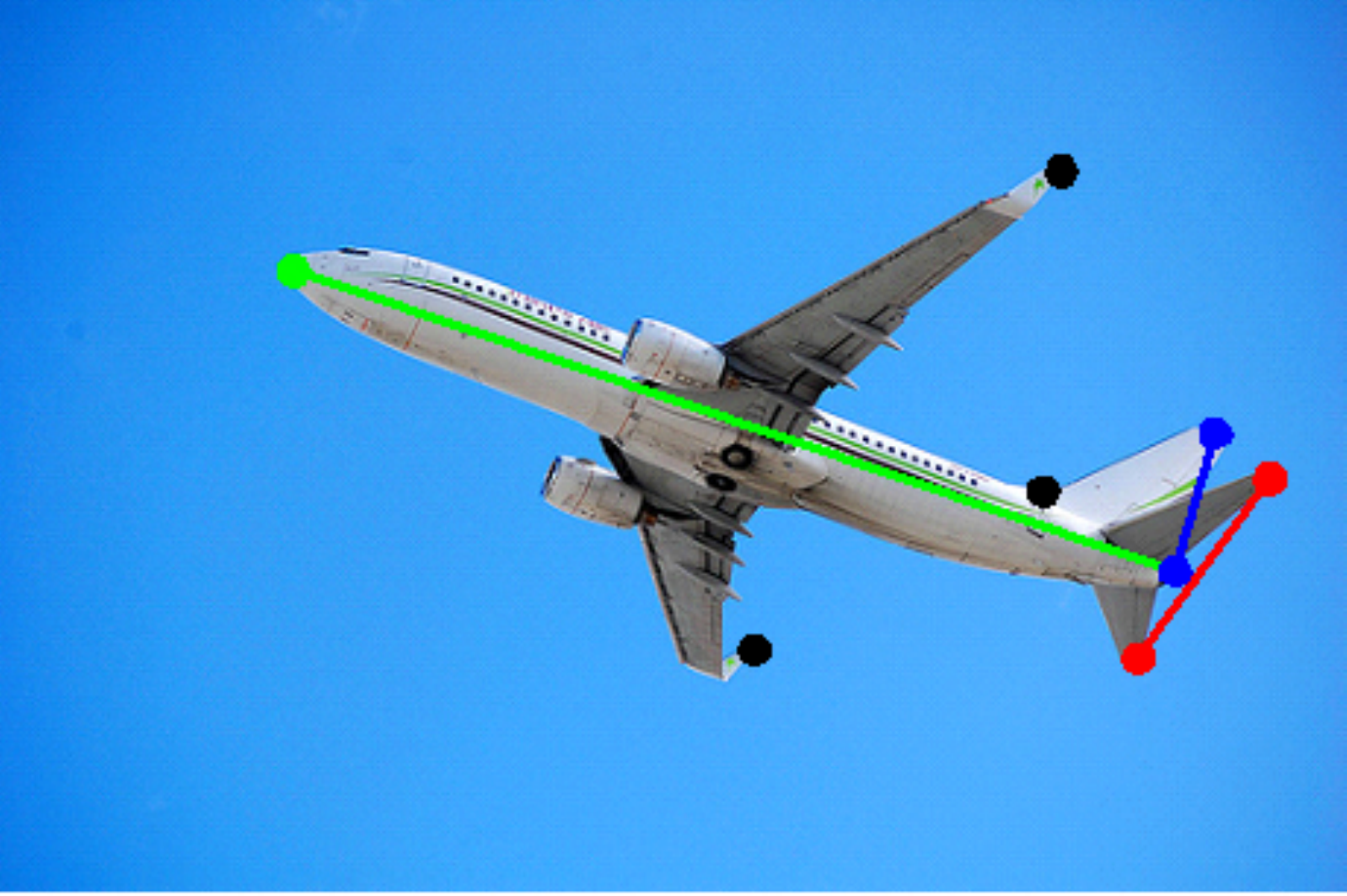}
\caption{Left Panel: Illustration of symmetry and Manhattan structure. The car has a bilateral symmetry with respect to the plane in red. There are three Manhattan axes. The first is normal to the symmetry plane of the car (\eg, from {\it left wheels} to {\it right wheels}). The second is from the front to the back of the car (\eg, from {\it back wheels} to {\it front wheels}) while the third is in the vertical direction. Right Panel: Illustration of the 3 Manhattan directions on a real aeroplane image, shown by Red, Green, Blue lines. These 3 Manhattan directions can be obtained directly from the labeled keypoints.}
\label{fig:symmetry_manhattan}
\end{figure}

Many objects, especially these made by humans, have intrinsic symmetry \cite{Rosen12,Hong04} and Manhattan properties (meaning that 3 perpendicular axes are inferable on the object \cite{Coughlan99,Coughlan03,Furukawa09}), such as cars, aeroplanes, see Fig~\ref{fig:symmetry_manhattan}. The purpose of this paper is to investigate the benefits of using symmetry and/or Manhattan constraints to estimate the 3D structures of objects from one or more images. As a key task in computer vision, numerous studies have been conducted on estimating the 3D shapes of objects from multiple images \cite{Hartley2004,Torresani03,Xiao04,Torresani08,Akhter2011,Gotardo11,Hamsici2012,Dai12,Dai14,Agudo14}. There is also a long history of research on the use of symmetry \cite{Gordon90,kontsevich93,Vetter94,Mukherjee95,Hong04,Thrun05,Li07,jiang2015polyhedral} and a growing body of work on Manhattan world \cite{Coughlan99,Coughlan03,Furukawa09}. There is, however, little work that combines these cues.

This paper aims at estimating the 3D structure of an object class, taking a single or multiple intra-class instances as input, \eg, different cars from various viewpoints. Following \cite{Vicente14,Kar15,Gao16}, we use 2D positions of keypoints  as input to estimate the 3D structure and the camera projection, leaving the detection of the 2D keypoints to methods such as \cite{chen2014articulated}. In this paper, different combinations of the three cues, \ie, symmetry, Manhattan and multiple images, are investigated and two algorithms/derivations are  proposed (assuming orthographic projection), \ie, single image reconstruction using both symmetry and Manhattan constraints, and multiple-image reconstruction using symmetry\footnote{We experimented with using Manhattan for the multiple-image case, but found it gave negligible improvement. Please see the derivations of using Manhattan for the multiple-image case in the supplementary material.}.

Specifically, we start with the single image reconstruction case, using the symmetry and Manhattan constraints, see Fig.~\ref{fig:symmetry_manhattan}. Our derivation is inspiring that a single image is sufficient for reconstruction when the symmetry and Manhattan axes can be inferred. Specifically, our analysis shows that using Manhattan alone is sufficient to recover the camera projection (up to several axis-sign ambiguities) by a single input image, then the 3D structure can be reconstructed uniquely by symmetry thereafter. We show these results on {\it aeroplane} in Pascal3D+. But we note that all the 3 Manhattan axes are hard to observe from a single image sometimes, particularly if the keypoints we rely on are occluded.

Hence, we extend the use of symmetry to the multiple-image case using structure from motion (SfM). The input is different intra-class object instances with various viewpoints. We formulate the problem in terms of energy minimization (\ie, MLE) of a probabilistic model, in which symmetry constraints are included. The energy also involves missing/latent variables for unobserved keypoints due to occlusion. These complications, and in particular symmetry, implies that we cannot directly apply singular value decomposition (SVD) (as in the previous SfM based work \cite{kontsevich87,Tomasi92}) to directly minimize the energy function. Instead we must rely on coordinate descent (or hard EM methods if we treat the missing points are latent variable) which risk getting stuck in local minimum. To address this issue we define a {\it surrogate energy function} which exploits symmetry, by grouping the keypoints into symmetric keypoint pairs, and assumes that the missing data are known (\eg, initialized by another process). We show that the surrogate energy can be decomposed into the sum of two independent energies, each of which can be minimized directly using SVD. This leads to a two-stage strategy where we first minimize the surrogate energy function as initialization for coordinate descent on the original energy function.

Recall that the classic SfM has a ``gauge freedom" \cite{morris2001gauge} because we can rotate the object and camera pose by equivalent amounts without altering the observations in the 2D images. This gauge freedom can be thought of as the freedom to {\it choose our coordinate system}. In this paper, we exploit this freedom to choose the symmetry axis to be in the $x$ axis (and the other two Manhattan axes to be in the $y$ and $z$ directions). 
In the following, we group keypoints into keypoint pairs and use a superscript $\dag$ to denote symmetry, \eg, $Y$ and $Y^{\dag}$ are symmetric keypoint pairs. 

The rest of the paper is organized as follows: firstly, we review related work in Section 2. In Section 3, we describe the common experiment design for all our methods. Then the mathematical details and evaluations on the single image reconstruction are given in Section 4, followed by the derivation and experiments on the multiple-image case, \ie, the  symmetric rigid structure from motion (Sym-RSfM), in Section 5. Finally, we give our conclusions in Section 6.

\section{Related Works}
Symmetry has been studied in computer vision for several decades. For example, symmetry has been used as a cue in depth recovery \cite{Gordon90,kontsevich93,Mukherjee95} as well as for recognizing symmetric objects \cite{Vetter94}. Grossmann and Santos-Victor utilized various geometric clues, such as planarity, orthogonality, parallelism and symmetry, for 3D scene reconstruction \cite{grossmann2002maximum,grossmann2005least}, where the camera rotation matrix was pre-computed by vanishing points \cite{grossmann2002single}.  Recently, researchers applied symmetry to scene reconstruction \cite{Hong04}, and 3D mesh reconstruction with occlusion \cite{Thrun05}. In addition, symmetry, incorporated with planarity and compactness priors, has also been studied to reconstruct structures defined by 3D keypoints \cite{Li07}. By contrast, the Manhattan world assumption was developed originally for scenes \cite{Coughlan99,Coughlan03,Furukawa09}, where the authors assumed visual scenes were based on a Manhattan 3D grid which provided 3 perpendicular axis constaints. Both symmetry and Manhattan can be straightforwardly combined, and adapted to 3D object reconstruction, particularly for man made objects.

The estimation of 3D structure from multiple images is one of the most active research areas in computer vision.
Classic SfM for rigid objects built on matrix factorization methods \cite{kontsevich87,Tomasi92}. Then, more general non-rigid deformation was considered, and the rigid SfM in \cite{kontsevich87,Tomasi92} was extended to non-rigid case by Bregler \etal \cite{Bregler00}. Non-rigid SfM was shown to have ambiguities \cite{Xiao04} and various non-rigid SfM methods were proposed using priors on the non-rigid deformations \cite{Xiao04,Torresani08,Olsen08,Akhter08,Gotardo11,Akhter2011}. Gotardo and Martinez proposed a Column Space Fitting (CSF) method for rank-$r$ matrix factorization and applied it to SfM with smooth time-trajectories assumption \cite{Gotardo11}. A more general framework for rank-$r$ matrix factorization was proposed in \cite{hongsecrets}, containing the CSF method as a special case\footnote{However, the general framework in \cite{hongsecrets} cannot be used to SfM directly, because it did not constrain that all the keypoints within the same frame should have the same translation. Instead, \cite{hongsecrets} focused on better optimization of rank-$r$ matrix factorization and pursuing better runtime.}. More recently, it has been proved that the ambiguities in non-rigid SfM do not affect the estimated 3D structure, \cite{Akhter09} which leaded to  prior free matrix factorization methods \cite{Dai12,Dai14}.

SfM methods have been used for category-specific object reconstruction, \eg, estimating the structure of \emph{cars} from images of different \emph{cars} under various viewing conditions \cite{Kar15,Vicente14}, but these did not exploit symmetry or Manhattan. We point out that in \cite{Ceylan14}, the repetition patterns have been incorporated into SfM for urban facades reconstruction, but \cite{Ceylan14} focused mainly on repetition detection and registration.

\section{Experimental Design}

This paper discusses 2 different scenarios to reconstruct the 3D structure: (i) reconstruction from a single image using symmetry and the Manhattan assumptions, (ii) reconstruction from multiple images using symmetric rigid SfM. The experiments are performed on Pascal3D+ dataset. This contains object categories such as \emph{aeroplane} and \emph{car}. These object categories are sub-divided into subtypes, such as \emph{sedan} car. For each object subtype, we estimate an 3D structure and the viewpoints of all the within-subtype instances. The 3D structure is specified by the 3D keypoints in Pascal3D+ \cite{Xiang14} and the corresponding keypoints in the 2D images are from Berkeley \cite{Bourdev10}. These are the same experimental settings as used in \cite{Kar15,Gao16}.

For evaluation we report the rotation error $e_R$ and the shape error $e_S$, as in \cite{Dai12,Dai14,Akhter08,Gotardo11,Gao16}. The 3D groundtruth and our estimates may have different scales, so we normalize them before evaluation. For each shape $S_n$ we use its standard deviations in $X, Y, Z$ coordinates $\sigma_n^x, \sigma_n^y, \sigma_n^z$ for normalization: $S_n^{\text{norm}} =  3 S_n/(\sigma_n^x + \sigma_n^y + \sigma_n^z)$. To deal with the rotation ambiguity between the 3D groundtruth and our result, we use the Procrustes method \cite{Schonemann66}  to align them. Assuming we have $2P$ keypoints, \ie, $P$ keypoint pairs, the rotation error $e_R$ and the shape error $e_S$ are calculated as:
\begin{align}
&e_R = \frac{1}{N} \sum_{n=1}^{N} ||R_n^{\text{aligned}} - R_n^* ||_F, \nonumber \\
&e_S = \frac{1}{2NP} \sum_{n=1}^{N} \sum_{p = 1}^{2P} ||S_{n,p}^{\text{norm aligned}} - S_{n,p}^{\text{norm}*}||_F,
\label{eq:Error}
\end{align}
where $R_n^{\text{aligned}}$ and $R_n^*$ are the recovered and the groundtruth camera projection matrix for image $n$. $S_{n,p}^{\text{norm aligned}}$ and $S_{n,p}^{\text{norm}*}$ are the normalized estimated structure and the normalized groundtruth structure for the $p$'th point of image $n$. These are aligned by the Procrustes method \cite{Schonemann66}.

\section{3D Reconstruction of A Single Image}

In this section, we describe how to reconstruct the 3D structure of an object from a single image using its symmetry and Manhattan properties. Theoretical analysis shows that this can be done with a bilateral symmetry and three Manhattan axes, but the estimation is ambiguous if less than three Manhattan axes are visible. Specifically, the three Manhattan constraints alone are sufficient to determine the camera projection up to sign ambiguities (\eg, we cannot distinguish between front-to-back and back-to-front directions). Then, the symmetry property is sufficient to estimate the 3D structure uniquely thereafter.

Let $Y, Y^{\dag} \in \mathbb{R}^{2 \times P}$ be the observed 2D coordinates of the $P$ symmetric pairs, then the orthographic projection implies:
\begin{equation}
Y = RS, \qquad Y^{\dag} = RS^{\dag}, \label{SingleImg}
\end{equation}
where $S, S^{\dag} \in \mathbb{R}^{3 \times P}$ are the 3D structure and $R \in \mathbb{R}^{2 \times 3}$ is the camera projection matrix. We have eliminated translation by centralizing the 2D keypoints.

\begin{remark}
We first estimate the camera projection matrix using the Manhattan constraints. Each Manhattan axis gives us one constraint on the camera projection. Hence, three axes give us the well defined linear equations to estimate the camera projection. We now describe this in detail.
\end{remark}

Consider a single Manhattan axis specified by 3D points $S_a$ and $S_b$. Without loss of generality, assume that these points are along the $x$-axis, \ie, $S_a - S_b = [x, 0, 0]^T$. It follows from the orthographic projection that:
\begin{equation}
Y_a - Y_b = R(S_a - S_b) = 
\begin{bmatrix}
r_{11}, & r_{12}, & r_{13} \\
r_{21}, & r_{22}, & r_{23}
\end{bmatrix}
\begin{bmatrix}
x \\ 0 \\ 0
\end{bmatrix}
=
\begin{bmatrix}
r_{11} x \\
r_{21} x
\end{bmatrix}. \label{x-axis}
\end{equation}
where $R = \begin{bmatrix} r_{11}, & r_{12}, & r_{13} \\ r_{21}, & r_{22}, & r_{23} \end{bmatrix}$. Setting $Y_a = [y_a^1, y_a^2]^T$, $Y_b = [y_b^1, y_b^2]^T$, yields
 $r_{21} / r_{11} = (y_a^2 - y_b^2) / (y_a^1 - y_b^1)$. With other Manhattan axes, \eg, if $S_c, S_d$ are along the $y$-axis, $S_e, S_f$ are along the $z$-axis, we can get similar constraints.

Let  $\mu_1 = r_{21}/r_{11}$, $\mu_2 = r_{22}/r_{12}$, $\mu_3 = r_{23}/r_{13}$, we have:
\begin{align} \mu _1 = (y_a^2 - y_b^2)/(y_a^1 - y_b^1), \nonumber \\ \mu _2 = (y_c^2 - y_d^2)/(y_c^1 - y_d^1), \nonumber \\ \mu _3 = (y_e^2 - y_f^2)/(y_e^1 - y_f^1),\label{eq:alan}\end{align}

Now, consider the orthogonality constraint on $R$, \ie, $RR^T = I$, which implies:
\begin{align}
& r_{11}^2 + r_{12}^2 + r_{13}^2 = 1, \nonumber \\ & r_{21}^2 + r_{22}^2 + r_{23}^2 = 1, \nonumber \\
& r_{11}r_{21} + r_{12}r_{22} + r_{13}r_{23} = 0.
\end{align}

Replacing $r_{21}, r_{22}, r_{23}$ by the known values $\mu_1, \mu_2, \mu_3$ of Eq. \eqref{eq:alan} indicates the following linear equations: 
\begin{equation}
\begin{bmatrix}
1, & 1, & 1 \\
\mu_1^2, & \mu_2^2, & \mu_3^2 \\
\mu_1, & \mu_2, & \mu_3
\end{bmatrix}
\begin{bmatrix}
r_{11}^2 \\ r_{12}^2 \\ r_{13}^2
\end{bmatrix}
=
\begin{bmatrix}
1 \\ 1 \\ 0
\end{bmatrix}
\end{equation}

These equations can be solved for the unknowns $r_{11}^2$, $r_{12}^2$ and $r_{13}^2$ provided the coefficient matrix (above) is invertible (\ie, has  full rank). This requires that
$(\mu_1 - \mu_2)(\mu_2 - \mu_3)(\mu_3 - \mu_1) \neq 0$. Because $\mu_1, \mu_2, \mu_3$ are  the slopes of the projected Manhattan axes in 3D space, this constraint is violated only in the very special case when the camera principal axis and two Manhattan axes are in the same plane.

Note that there are sign ambiguities for solving $r_{11}, r_{12}, r_{13}$ from $r_{11}^2, r_{12}^2, r_{13}^2$. But these ambiguities do not affect the estimation of the 3D shape, because they are just choices of the coordinate system. Next we can calculate  $r_{21}, r_{22}, r_{23}$ directly based on $r_{11}, r_{12}, r_{13}$ and $\mu_1, \mu_2, \mu_3$. This recovers the projection matrix.

\begin{remark}
We have shown that the camera projection matrix $R$ can be recovered if the three Manhattan axes are known. Next we show that the 3D structure can be estimated using symmetry (provided the projection $R$ is recovered).
\end{remark}

Assume, without loss of generality, the object is along the $x$-axis. Let $Y \in \mathbb{R}^{2 \times P}$ and $Y^{\dag} \in \mathbb{R}^{2 \times P}$ be the $P$ symmetric pairs, $S \in \mathbb{R}^{3 \times P}$ and $S^{\dag} \in \mathbb{R}^{3 \times P}$ be the corresponding $P$ symmetric pairs in the 3D space. Note that for the $p$'th point pair in the 3D space, we have $S_p = [x_p, y_p, z_p]^T$ and $S_p^{\dag} = [-x_p, y_p, z_p]$. Thus, we can re-express the camera projection in Eq.~\ref{SingleImg} by:
\begin{align}
L &= \frac{Y - Y^{\dag}}{2} = R \begin{bmatrix} x_1, \   ..., \ x_P \\ 0, \   ..., \ 0 \\ 0, \   ..., \ 0 \end{bmatrix} =
\begin{bmatrix}
r_{11} x_1, \   ..., \  r_{11} x_P \\
r_{21} x_1, \   ..., \  r_{21} x_P
\end{bmatrix},
\label{SS1}\\
M &= \frac{Y + Y^{\dag}}{2} = R \begin{bmatrix} 0, \   ..., \ 0 \\ y_1, \   ..., \ y_P \\ z_1, \   ..., \ z_P \end{bmatrix} \nonumber \\
& = \begin{bmatrix}
r_{12} y_1 + r_{13} z_1, \   ..., \ r_{12} y_P + r_{13} z_P \\
r_{22} y_1 + r_{23} z_1, \   ..., \ r_{22} y_P + r_{23} z_P
\end{bmatrix} \label{SS2}.
\end{align}
Finally, we can solve Eqs. \eqref{SS1} and \eqref{SS2} to estimate the components $(x_p,y_p,z_p)$ of all the points $S_p$ (since $R$ is known), and hence, recover the 3D structure. Observe that we have only just enough equations to solve $(y_p,z_p)$ uniquely. On the other hand, the $x_p$ is over-determined due to symmetry. We also note that the problem is ill-posed if we do not exploit symmetry, \ie, it involves inverting a $2 \times 3$ projection matrix $R$ if not exploiting symmetry.

\subsection{Experiments on 3D Reconstruction Using A Single Image}
We use \emph{aeroplane}s for this experiment, because the 3 Manhattan directions (\eg, left wing $\rightarrow$ right wing, nose $\rightarrow$ tail and top rudder $\rightarrow$ bottom rudder) can be obtained directly on aeroplanes, see Fig. \ref{fig:symmetry_manhattan}. Also aeroplanes are generally far away from the camera, implying orthographic projection is a good approximation.

We selected 42 images with clear 3 Manhattan directions and with no occluded keypoints from the \emph{aeroplane} category of Pascal3D+ dataset, and evaluated the results by the Rotation Error and Shape Error (Eq. \eqref{eq:Error}). The shape error is obtained by comparing the reconstructed structure with their subtype groundtruth model of Pascal3D+ \cite{Xiang14}.

\begin{figure}[t]
\centering
\includegraphics[width=\linewidth]{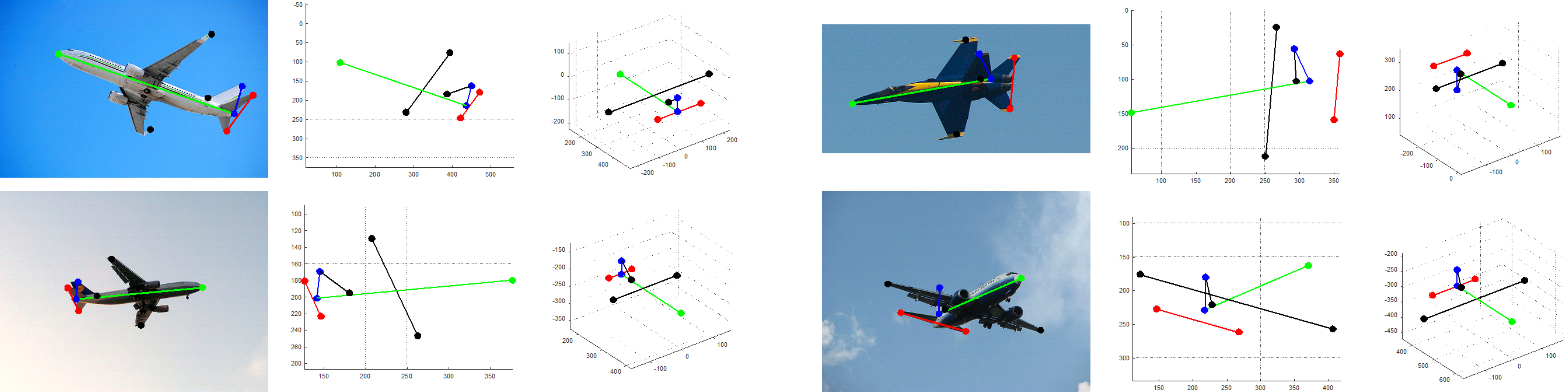}
\caption{Illustration of the reconstruction results for \emph{aeroplane} using the symmetry and Manhattan constraints on a single image. For each subfigure triplet, the first subfigure is the 2D image with input keypoints, the second and third subfigures are the 3D structure from the original and rectified viewpoints. The gauge freedom of sign ambiguities can also be observed by comparing the rectified 3D reconstructions, \ie, the third subfigures. The Red, Green, Blue lines represent the three Manhattan directions we used.}
\label{single_illust}
\end{figure}

The \emph{average rotation and shape errors} for aeroplane using the Manhattan and symmetry constraints on the single image case are \emph{0.3210} and \emph{0.6047}, respectively. These results show that using the symmetry and Manhattan properties alone can give good results for the single image reconstruction. Indeed, the performance is better than some of the structure from motion (SfM) methods which use multiple images, see Tables \ref{table:Rot_Results} and \ref{table:Shp_Results} (on Page 8). But this is not a fair comparison, because these 42 images are selected to ensure that all the Manhattan axes are visible, while the SfM methods have been evaluated on all the aeroplane images. Some reconstruction results are illustrated in Fig. \ref{single_illust}.

\section{Symmetric Rigid Structure from Motion}

This section describes symmetric rigid structure from motion  (Sym-RSfM). We start by defining the a full energy function for the problem, see Section~\ref{sec:original}, and the coordinate descent algorithm to minimize it, see Section~\ref{sec:coordinatedescent}. Then, the missing points are initialized in Section~\ref{sec:missingData}. After that, we describe the surrogate energy and the algorithm to minimize it, see Section~\ref{sec:surrogate}, which serves to initialize coordinate descent on the full energy function.

For consistency with our baseline methods \cite{Tomasi92}, we assume  orthographic projection and the keypoints are centralized without translation for the Sym-RSfM. Note that due to the iterative estimation and recovery of the occluded data, the translation has to be re-estimated to re-centralize the data during each iteration. The update of the translation is straightforward and will be given in Section~\ref{sec:coordinatedescent}.

\subsection{Problem Formulation: The Full Energy \label{sec:original}}
We extend the image formation model from a single image to multiple images indexed by $n=1,...,N$. The keypoints are grouped into $P$ keypoint pairs, which are symmetric across the $x$-axis. We introduce noise in the image formation, and the translation is removed as discussed above.

Then, by assuming that the 3D structure $S$ is symmetric along the $x$-axis, the 2D keypoint pairs for image $n$, \ie, $Y_n \in \mathbb{R}^{2 \times P}$ and $Y^{\dag}_n \in \mathbb{R}^{2 \times P}$, are given by:
\begin{align}
Y_n = R_n S + N_n \ &\Rightarrow \ P(Y_n| R_n, S) \sim \mathcal{N}(R_n S, 1) \nonumber \\
Y^{\dag}_n = R_n \a S + N_n \ &\Rightarrow \ P(Y_n^{\dag}| R_n, S) \sim \mathcal{N}(R_n \a S, 1) \label{rigid_model}
\end{align}
where $R_n \in \mathbb{R}^{2 \times 3}$ is the projection matrix, $S \in \mathbb{R}^{3 \times P}$ is the 3D structure, and $\mathcal{A}$ is a matrix operator $\mathcal{A} = \text{diag}([-1, 1, 1])$, by which $\a S$ changes the sign of the first row of $S$, making $S^{\dag} = \a S$. $N_n$ is zero mean Gaussian noise with unit variance $N_n \sim \mathcal{N}(0,1)$\footnote{Note that we experimented with another model which treated the variance as an unknown parameter $\sigma^2_n$ and estimated it during optimization, but found this made negligible difference to the experimental results.}.

Observe that the noise is independent for all keypoints and for all images. Hence, the 2D keypoints are independent when conditioned on the 3D structure $S$ and the camera projections $R_n$. Therefore, the problem is formulated in terms of
minimizing the following energy function with unknown $R_n, S$:
\begin{align}
\mathcal{Q}(R_n, S) &=  -\sum_n \ln P(Y_n, Y^{\dag}_n | R_n, S) \nonumber \\ & = -\sum_n \left( \ln P(Y_n| R_n, S) - \ln P(Y^{\dag}_n | R_n, S) \right) \nonumber \\
&\sim \sum_n ||Y_n - R_n S||^2_2 + \sum_n||Y^{\dag}_n - R_n \a S ||^2_2. \label{RigidEner0}
\end{align}

This formulation relates to the classic structure from motion problems \cite{kontsevich87,Tomasi92}, but those classic methods do not impose  symmetry,  and therefore, they have only the first term in Eq. \eqref{RigidEner0}, \ie, $\mathcal{Q}(R_n, S) = \sum_n||Y_n - R_n S||^2_2$. If all the data is fully observed then $S$ is of rank 3 (assuming the 3D points do not lie in a plane or a line) and so the energy can be minimized by first stacking the keypoints for all the images together, \ie, $\mathbf{Y} = [Y_1^T, ..., Y_N^T]^T \in \mathbb{R}^{2N \times P}$, then applying SVD to $\mathbf{Y}$. The solution is unique up to some coordinate transformations, \ie, the rotation ambiguities \cite{kontsevich87,Tomasi92}.

\begin{remark}
Our problem is more complex than the classic rigid SfM because of two confounding issues: (i) some keypoints will be unobserved in each image, (ii) we cannot directly solve Eq. \eqref{RigidEner0} by SVD because it consists of two energy terms which are not independent (even if all data is observed). Hence, we first formulate a full energy function with missing points, where the missing points are initialized in Section~\ref{sec:missingData}. After that, in Section~\ref{sec:surrogate}, a surrogate energy is defined, which exploits symmetry and can be minimized by SVD, and therefore, can be used for initializing camera projection and the 3D structure.
\end{remark}

To deal with unobserved keypoints we divide them into {\it visible sets} ${\it VS}, {\it VS}^{\dag}$and {\it invisible sets} ${\it IVS}, {\it IVS}^{\dag}$. Then the {\it full energy function} can be formulated as:
\vspace{-1mm}
{\small
\begin{align}
&\mathcal{Q}(\mathbf{R}, S, \{Y_{n,p}, (n,p) \in {\it IVS}\}, \{Y_{n,p}^{\dag}, (n,p) \in {\it IVS}^{\dag}\}) \nonumber \\
=&  \sum_{(n,p) \in {\it VS}} ||Y_{n,p} - R_n S_{n, p}||^2_2 + \sum_{(n,p) \in {\it VS}^{\dag}}||Y_{n,p}^{\dag} - R_n \a S_{n, p} ||^2_2 + \nonumber \\
& \sum_{(n,p) \in I{\it VS}} ||Y_{n,p} - R_n S_{n, p}||^2_2 + \sum_{(n,p) \in {\it IVS}^{\dag}}||Y_{n,p}^{\dag} - R_n \a S_{n, p} ||^2_2. \label{RigidEner2}
\end{align}
}Here $\{Y_{n,p}, (n,p) \in {\it IVS}\}$, $\{Y_{n,p}^{\dag}, (n,p) \in {\it IVS}^{\dag}\}$ are the missing points. 

Note that it is easy to add the Manhattan constraints in Eq. \eqref{RigidEner2} as a regularization term for the Sym-RSfM on multiple images. But we found no significant improvement in our experiments when we used Manhattan, perhaps because it was not needed due to the extra images available as input. Please see the supplementary materials.

\subsection{Optimization of The Full Energy Function \label{sec:coordinatedescent}}

We now define a \emph{coordinate descent} algorithm to estimate the 3D structure, the camera projection, and the missing data. This algorithm is not guaranteed to converge to the global minimum, so it will be initialized using the surrogate energy function, described in Section~\ref{sec:surrogate}.

We use a \emph{coordinate descent} method to optimize Eq. \eqref{RigidEner2} by updating $R_n, S$ and the missing points $\{Y_{n,p}, (n,p) \in IVS\}, \{Y_{n,p}^{\dag}, (n,p) \in IVS^{\dag}\}$ iteratively. Note that the energy in Eq. \eqref{RigidEner2} \emph{w.r.t} $R_n, S$, \ie, when the missing points are fixed, is given in Eq. \eqref{RigidEner0}.

Firstly, we vectorize Eq. \eqref{RigidEner0} and $S$ to update it in matrix form by $\mathbb{S}$:
\begin{equation}
\scalebox{0.8}{$
\mathbb{S} = \left( \sum_{n=1}^{N} (G_n^TG_n + \a_P^T G_n^T G_n \a_P) \right)^{-1} \left(\sum_{n=1}^{N} ( G_n^T \mathbb{Y}_n +        \a_P^T G_n^T \mathbb{Y}_n^{\dag} ) \right).  \label{Rigid-S1}
$}
\end{equation}
where $\mathbb{S} \in \mathbb{R}^{3P \times 1}, \mathbb{Y}_n \in \mathbb{R}^{2P \times 1}, \mathbb{Y}_n^{\dag} \in \mathbb{R}^{2P \times 1}$ are vectorized $S, Y_n, Y_n^{\dag}$, respectively. $G_n = I_{P} \otimes R_n$ and $\a_P = I_{P} \otimes \a$. $I_{P} \in \mathbb{R}^{P \times P}$ is an identity matrix.

Each $R_n$ is updated under the nonlinear orthogonality constraints $R_n R_n^T = I$ similar to the idea in EM-PPCA \cite{Torresani08}: we first parameterize $R_n$ to a full $3 \times 3$ rotation matrix $Q$ and update $Q$ by its rotation increment. Please refer to the supplementary materials for the details.

\begin{algorithm}[t]
\SetNlSkip{0.5em}
\SetInd{0.5em}{1em}
\caption{Optimization of the full energy Eq. \eqref{RigidEner2}.}
\label{algo1}
\KwIn{The stacked keypoint sets (for all the $N$ images) $\mathbf{Y}$ and $\mathbf{Y}^{\dag}$ with occluded points, in which each occluded point is set to $\mathbf{0}$ initially.
}
\KwOut{The camera projection matrix $R_n$ for each image, the 3D structure $S$, and the keypoints with recovered occlusions $(\mathbf{Y})^t$ and $(\mathbf{Y}^{\dag})^t$.}
Initialize the occluded points by Algorithm \ref{algo2}. \\
Initialize the each camera projection $R_n$ and the 3D structure $S$ by Algorithm \ref{algo3}. \\
\Repeat{Eq. \eqref{RigidEner2} converge}{
  Update $S$ by Eq. \eqref{Rigid-S1} and update each $R_n$ (see the supplementary materials).  
  \\
  Calculate the occluded points by Eq. \eqref{Rigid-OP}, and update them in $Y_n, Y_n^{\dag}$. \\
  Centralize the $Y_n, Y_n^{\dag}$ by Eq. \eqref{Rigid-centralize}.
}
\end{algorithm}

From Eq. \eqref{RigidEner2}, the occluded points of $\mathbf{Y}$ and $\mathbf{Y}^{\dag}$ (\ie, the $p$-th point $Y_{n,p}$ and $Y_{n,p}^{\dag}$) can be updated by minimizing: $\mathcal{Q}(Y_{n,p}, Y_{n,p}^{\dag}) = \sum_{(n,p) \in IVS} ||Y_{n,p} - R_n S_p||^2_2 + \sum_{(n,p) \in IVS^{\dag}}||Y_{n,p}^{\dag} - R_n \a S_p ||^2_2$, which implies the update rule for the missing points:
\begin{equation}
Y_{n,p} = R_n S_p, \qquad Y_{n,p}^{\dag} = R_n \a S_p, \label{Rigid-OP}
\end{equation}
where $(n,p) \in IVS$.

Note that we do not model the translation explicitly for the sake of consistency with the baseline method \cite{Tomasi92}, where the translation is assumed to be eliminated by centralizing the data. However, since the occluded points have been updated iteratively in our method, we have to re-estimate the translation and re-centralize the data during each iteration. This can be done by:
\begin{align}
  &Y_n \gets Y_n - \mathbf{1}_{2P}^T \otimes t_n, \qquad Y_n^{\dag} \gets Y_n^{\dag} - \mathbf{1}_{2P}^T \otimes t_n,  \nonumber \\
  &t_n = \sum_p (Y_{n,p} - R_n S_p + Y_{n,p}^{\dag} - R_n \mathcal{A} S_p).  \label{Rigid-centralize}
\end{align}

The algorithm to optimize Eq. \eqref{RigidEner2} is summarized in Algorithm \ref{algo1}, in which the initialization of the missing points, the 3D structure and the camera projection, \ie, Algorithms \ref{algo2} and \ref{algo3}, will be discussed in the following sections.

\subsection{Initialization of The Missing Data \label{sec:missingData}}
In this section, the missing data is initialized by the whole input data ignoring symmetry. This will be used both for coordinate descent of the full energy and for applying singular value decomposition to the surrogate energy.

Let $\mathbf{Y} = [Y_1^T, ..., Y_N^T]^T, \mathbf{Y}^{\dag} = [(Y_1^{\dag})^T, ..., (Y_N^{\dag})^T]^T \in \mathbb{R}^{2N \times P}$ are the stacked keypoints for all the images, and $\mathbf{R} = [R_1^T, ...R_N^T]^T \in \mathbb{R}^{2N \times 3}$ are the stacked camera projection. Thus, we have $\mathbf{Y}^{\text{All}} = [\mathbf{Y}$, $\mathbf{Y}^{\dag}] = \mathbf{R} [S, \a S]$. It implies that $\mathbf{Y}^{\text{All}}$ has the same rank, namely 3, with $\mathbf{R} [S, \a S]$ given all the points of $[S, \a S]$ do not lie on a plane or a line. Therefore, rank 3 recovery can be used to initialize the missing points. Also, the same centralization as in the previous section has to be done after each iteration of the missing points, so as to eliminate the translations.

The occlusions initialization is shown in Algorithm \ref{algo2}.

\subsection{The Surrogate Energy: Initialization of Camera Projection and 3D Structure
\label{sec:surrogate}}

\begin{remark}
We now define a surrogate energy function that exploits the symmetry constraints, which enables us to decompose the energy into two independent terms and leads to an efficient minimization algorithm using SVD.
\end{remark}

To construct the surrogate energy, we first change the coordinates to exploit symmetry, so that the problem breaks down into two independent energy terms. Since $S$ and $S^{\dag}$ are symmetric along x-axis, we can decompose $S$ by:
{\small
\begin{align}
& \mathbf{L} = \frac{\mathbf{Y} - \mathbf{Y}^{\dag}}{2} = \mathbf{R} \begin{bmatrix} x_1, &..., &x_P \\ 0, &..., &0 \\ 0, &..., &0 \end{bmatrix} = \mathbf{R}^1 S_x, \quad \nonumber \\
& \mathbf{M} = \frac{\mathbf{Y} + \mathbf{Y}^{\dag}}{2} = \mathbf{R} \begin{bmatrix} 0, &..., &0 \\ y_1, &..., &y_P \\ z_1, &..., &z_P \end{bmatrix} = \mathbf{R}^2 S_{yz} \label{SRSfM2},
\end{align}
}where $\mathbf{R}^1 \in \mathbb{R}^{2N \times 1}, \mathbf{R}^2 \in \mathbb{R}^{2N \times 2}$ are the first single column and second-third double columns of $\mathbf{R}$, $S_x \in \mathbb{R}^{1 \times P}, S_{yz} \in \mathbb{R}^{2 \times P}$ are the first single row and second-third double rows of $S$, respectively. Equation \eqref{SRSfM2} gives us the energy function on $\mathbf{R}, S$ to replace Eq. \eqref{RigidEner0} into:
\begin{equation}
Q(\mathbf{R}, S) = ||\mathbf{L} - \mathbf{R}^1 S_x||_2^2 + ||\mathbf{M} - \mathbf{R}^2 S_{yz}||_2^2. \label{SVD}
\end{equation}

This is essentially changing the coordinate system by rotating $\mathbf{Y}, \mathbf{Y}^T$ with 45$^{\circ}$ (except a scale factor of $\sqrt{2}$).

\begin{algorithm}[t]
\SetNlSkip{0.5em}
\SetInd{0.5em}{1em}
\caption{The initialization of the occluded points.}
\label{algo2}
\KwIn{The stacked keypoint sets (for all the $N$ images) $\mathbf{Y}$ and $\mathbf{Y}^{\dag}$ with occluded points, in which each occluded point is set to $\mathbf{0}$ initially. The number of iterations $T$ (default 10).}
\KwOut{The keypoints with initially recovered occlusions $(\mathbf{Y})^t$ and $(\mathbf{Y}^{\dag})^t$.}
Set $t = 0$, initialize the occluded points ignoring symmetry by: \\
\While{$t < T$}{
  Centralize $\mathbf{Y}^{\text{All}} = [(\mathbf{Y})^t$, $(\mathbf{Y}^{\dag})^t]$ by Eq. \eqref{Rigid-centralize}. \\
  Do SVD on $\mathbf{Y}^{\text{All}}$ ignoring the symmetry, \ie, $[\mathbf{A}, \Sigma, \mathbf{B}] = \text{SVD}\left(\mathbf{Y}^{\text{All}}\right)$. \\
  Use the first 3 component of $\Sigma$ to reconstruct the keypoints $(\mathbf{Y}^{\text{All}})^{\text{new}}$. \\
  Replace the occluded points in $(\mathbf{Y})^t$, $(\mathbf{Y}^{\dag})^t$ by these in $(\mathbf{Y}^{\text{All}})^{\text{new}}$ and set $t \leftarrow t+1$.
}
\end{algorithm}

\begin{remark}
We have decomposed the energy into two independent terms, and therefore, they can be solved separately by SVD up to some ambiguities. Then we will combine them to study and resolve the ambiguities. Note that we assume the occluded keypoints are replaced by the initialization described in the previous section.
\end{remark}

Equation \eqref{SVD} implies that we can estimate $\mathbf{R}^1, S_x$ and $\mathbf{R}^2, S_{yz}$ by matrix factorization on $\mathbf{L}$ and $\mathbf{M}$ independently up to ambiguities. Then we combine them to remove this ambiguity by exploiting the orthogonality constraints on each $R_n$: \ie, $R_n R_n^T = I$. Applying SVD to $\mathbf{L}$ and $\mathbf{M}$ gives us estimates, \ie, $(\mathbf{\hat{R}}^1, \hat{S}_x)$ of $(\mathbf{R}^1, S_x)$, and $(\mathbf{\hat{R}}^2, \hat{S}_{yz})$ of $(\mathbf{R}^2, S_{yz})$, up to ambiguities $\lambda$ and $B$:
\begin{equation} \mathbf{L} = \mathbf{R}^1 S_x = \mathbf{\hat{R}}^1 \lambda \lambda^{-1} \hat{S}_x, \quad \mathbf{M} = \mathbf{R}^2 S_{yz} = \mathbf{\hat{R}}^2 B B^{-1} \hat{S}_{yz}, \label{ambi}\end{equation}
here $\mathbf{R}^1$ and $\mathbf{R}^2$ are the decomposition of the true projection matrix $\mathbf{R}$, \ie, $\mathbf{R} = [\mathbf{R}^1, \mathbf{R}^2]$, and $\mathbf{\hat{R}}^1$ and $\mathbf{\hat{R}}^2$ are the output estimates from SVD. Equation \eqref{ambi}  shows that there is a scale ambiguity $\lambda$ between $\mathbf{\hat{R}}^1$ and $\mathbf{R}^1$, and a 2-by-2 matrix ambiguity $B \in \mathbb{R}^{2 \times 2}$ between $\mathbf{\hat{R}}^2$ and $\mathbf{R}^2$.

\begin{remark}
Next we show how to resolve the ambiguities $\lambda$ and $B$. This is done by using the \emph{orthogonality constraints}, namely $R_n R_n^T = I$.
\end{remark}

Observe from Eqs. \eqref{ambi}  that the ambiguities (\ie, $\lambda$ and $B$) are the same for the projection matrices of all the images. In the following derivation, we analyze the ambiguity for the $n$'th image, \ie, projection matrix $R_n$.

Using Eqs. \eqref{ambi}, the true $R_n$ can be represented by:
\begin{equation}
R_n = [R_n^1, R_n^2] = [\hat{R}^1_n, \hat{R}^2_n]
\begin{bmatrix}
\lambda, & \mathbf{0} \\
\mathbf{0}, & B
\end{bmatrix} \label{Rotation} \\
=
\hat{R}_n
\begin{bmatrix}
\lambda, & \mathbf{0} \\
\mathbf{0}, & B
\end{bmatrix}\end{equation}
where $R_n^1 \in \mathbb{R}^{2 \times 1}$ and $R_n^2 \in \mathbb{R}^{2 \times 2}$ are the first single
column and second-third double columns of the true projection matrix $R_n$. $\hat{R}^1_n \in \mathbb{R}^{2 \times 1}$ and $\hat{R}_n^2 \in \mathbb{R}^{2 \times 2}$ are the initial estimation of $R_n^1$ and $R_n^2$ from the matrix factorization.

Let $\hat{R}_n = [\hat{R}^1_n, \hat{R}^2_n] = \begin{bmatrix} \hat{r}^{1,1}_n, & \hat{r}^{1,2:3}_n \\ \hat{r}^{2,1}_n, & \hat{r}^{2,2:3}_n \end{bmatrix} \in \mathbb{R}^{2 \times 3}$, imposing the orthogonality constraints $R_n R_n^T = I$ using Eq. \eqref{Rotation} gives:
{\small
\begin{align}
& R_n R_n^T = \hat{R}_n \begin{bmatrix}
\lambda^2, & \mathbf{0} \\
\mathbf{0}, & BB^T
\end{bmatrix}
\hat{R}_n^T \nonumber \\
=& \begin{bmatrix}
\hat{r}^{1,1}_n, & \hat{r}^{1,2:3}_n \\
\hat{r}^{2,1}_n, & \hat{r}^{2,2:3}_n
\end{bmatrix}
\begin{bmatrix}
\lambda^2, & \mathbf{0} \\
\mathbf{0}, & BB^T
\end{bmatrix}
\begin{bmatrix}
\hat{r}^{1,1}_n, & \hat{r}^{1,2:3}_n \\
\hat{r}^{2,1}_n, & \hat{r}^{2,2:3}_n
\end{bmatrix}^T = I  \label{R_orth}
\end{align}
}

Vectorizing $BB^T$ of Eq. \eqref{R_orth} using $\text{vec}(AXB^T) = (B \otimes A)\text{vec}(X)$, we can get the following linear equations:
\begin{equation}
\begin{bmatrix}
(\hat{r}^{1,1}_n)^2, & \hat{r}^{1,2:3}_n \otimes \hat{r}^{1,2:3}_n \\
(\hat{r}^{2,1}_n)^2, & \hat{r}^{2,2:3}_n \otimes \hat{r}^{2,2:3}_n \\
\hat{r}^{1,1}_n \hat{r}^{2,1}_n, & \hat{r}^{1,2:3}_n \otimes \hat{r}^{2,2:3}_n \\
\end{bmatrix}
\begin{bmatrix}
\lambda^2 \\
\text{vec}(BB^T)
\end{bmatrix}
=
\begin{bmatrix}
1 \\
1 \\
0
\end{bmatrix} .
\label{Rigid_orth}
\end{equation}
Note that $BB^T$ is a symmetric matrix, the second and third elements of $\text{vec}(BB^T)$ are the same. Let $\text{vec}(BB^T) = [bb_1, bb_2, bb_2, bb_3]^T$, we can enforce the symmetriy of $B B^T$ by rewriting Eq. \eqref{Rigid_orth}:
\begin{equation}
\begin{bmatrix}
(\hat{r}^{1,1}_n)^2, & \hat{r}^{1,2:3}_n \otimes \hat{r}^{1,2:3}_n \\
(\hat{r}^{2,1}_n)^2, & \hat{r}^{2,2:3}_n \otimes \hat{r}^{2,2:3}_n \\
\hat{r}^{1,1}_n \hat{r}^{2,1}_n, & \hat{r}^{1,2:3}_n \otimes \hat{r}^{2,2:3}_n \\
\end{bmatrix}
\begin{bmatrix}
1 \ 0 \ 0 \ 0 \\
0 \ 1 \ 0 \ 0 \\
0 \ 0 \ 1 \ 0 \\
0 \ 0 \ 1 \ 0 \\
0 \ 0 \ 0 \ 1
\end{bmatrix}
\begin{bmatrix}
\lambda^2 \\
bb_1 \\
bb_2 \\
bb_3
\end{bmatrix}
= A_i \mathbf{x} =
\begin{bmatrix}
1 \\
1 \\
0
\end{bmatrix} ,
\label{Rigid_orth2}
\end{equation}
where the constant matrix of the left term is a matrix operator to sum the third and forth columns of the coefficient matrix with $r$'s, $\mathbf{x}$ is the unknown variables $[\lambda^2, bb_1, bb_2, bb_3]^T$, and $A_i \in \mathbb{R}^{3 \times 4}$ is the corresponding coefficients.

\begin{algorithm}[t]
\SetNlSkip{0.5em}
\SetInd{0.5em}{1em}
\caption{The initialization of the camera projection and structure.}
\label{algo3}
\KwIn{The keypoint sets $\mathbf{Y}$ and $\mathbf{Y}^{\dag}$ with initially recovered occluded points by Algorithm \ref{algo2}.}
\KwOut{The initialized camera projection $\mathbf{R}$ and the 3D structure $S$.}
Change the coordinates to decouple the symmetry constraints by Eq. \eqref{SRSfM2}. \\
Get $\mathbf{\hat{R}}^1, \mathbf{\hat{R}}^2, \hat{S}_x, \hat{S}_{yz}$ by SVD on $\mathbf{L}, \mathbf{M}$, \ie, Eq. \eqref{ambi}. \\
Solve the squared ambiguities $\lambda ^2$, $BB^T$ by Eq. \eqref{Rigid_orth2}. \\
Solve for $\lambda$ from $\lambda^2$, and $B$ from $BB^T$, up to sign and rotation ambiguities. \\
Obtain the initialized $\mathbf{R}$ and $S$ by Eq. \eqref{initial_RS}.
\end{algorithm}

Stacking all the $\hat{\mathbf{R}}$'s together, \ie, let $\mathbf{A} = [A_1^T, ..., A_N^T]^T \in \mathbb{R}^{3N \times 4}$ and $\mathbf{b} = \mathbf{1}_{N} \otimes [1,1,0]^T$, we have a over-determined equations for the unknown $\mathbf{x}$: $\mathbf{A}\mathbf{x} = \mathbf{b}$ (\ie, $3N$ equations for 4 unknowns), which can be solved efficiently by LSE: $\mathbf{x} = (\mathbf{A}^T\mathbf{A})^{-1}\mathbf{A}^T\mathbf{b}$.

\begin{remark}
The ambiguity of $\mathbf{R}^1$ and $\hat{\mathbf{R}}^1$, (\ie, in the symmetry direction), is just a sign change, which cause by calculating $\lambda$ from  $\lambda^2$. In other words, the symmetry direction can be fixed as $x$-axis in our coordinate system using the decomposition Eq. \eqref{SRSfM2}.
\end{remark}

After obtained $BB^T$, $B$ can be recovered up to a rotation ambiguity on $yz$-plane, which does not affect the reconstructed 3D structure (See the supplementary materials).

Given $\lambda, B, \hat{\mathbf{R}}, \hat{S}$, we can get the true $\mathbf{R}$ and $S$ by:
\begin{equation}
\mathbf{R} = \mathbf{\hat{R}}
\begin{bmatrix}
\lambda, & \mathbf{0} \\
\mathbf{0}, & B
\end{bmatrix}, \qquad
S =
\begin{bmatrix}
\lambda, & \mathbf{0} \\
\mathbf{0}, & B
\end{bmatrix}^{-1} \hat{S}.  \label{initial_RS}
\end{equation}

\begin{table*}[t!]
\fontsize{9pt}{0.9\baselineskip}\selectfont
\begin{tabular*}{\textwidth}{@{\extracolsep{\fill}} || l || c | c | c | c | c | c | c || c | c | c | c | c | c ||}
\hline
& \multicolumn{7}{c||}{\textbf{aeroplane}} & \multicolumn{6}{c||}{\textbf{bus}} \\
\hline
& I & II & III & IV & V & VI & VII &  I & II & III & IV & V & VI  \\
\hline
RSfM &  0.52 & 1.17 & 1.99 & 0.28 & 0.94 & 1.77 & 1.74 & 0.32 & 1.05 & 0.27 & 0.27 & 1.13 & 0.51 \\
CSF (S) & 0.92      & 0.70      & 0.87      & 0.83      & \underline{\textbf{0.89}}      & 0.96      & \underline{\textbf{1.02}} & 0.72      & 0.86      & 0.68      & 0.92      & 0.94      & 1.04      \\
CSF (R) & 0.93      & 0.82      & \underline{\textbf{0.80}}      & 0.91      & 0.99      & 1.02      & 1.37 & 0.69      & 0.88      & 0.81      & 0.93      & 0.88      & 1.08      \\
Sym-RSfM   & \underline{\textbf{0.13}} & \underline{\textbf{0.46}} & 2.00 & \underline{\textbf{0.17}} & 1.81 & \underline{\textbf{0.89}} & 1.69 & \underline{\textbf{0.19}} & \underline{\textbf{0.33}} & \underline{\textbf{0.03}} & \underline{\textbf{0.22}} & \underline{\textbf{0.58}} & \underline{\textbf{0.50}} \\
\hline
\end{tabular*}

\begin{tabular*}{\textwidth}{@{\extracolsep{\fill}} || l || c | c | c | c | c | c | c | c | c | c || c | c ||}
\hline
& \multicolumn{10}{c||}{\textbf{car}} & \multicolumn{2}{c||}{\textbf{sofa}} \\
\hline
& I & II & III & IV & V & VI & VII & VIII & IX & X  & I & II  \\
\hline
RSfM &  0.58 & 0.71 & 0.54 & 1.10 & 0.67 & 1.51 & 0.67 & 1.41 & 0.97 & 0.37 & 1.18 & 0.75 \\
CSF (S) & 0.95      & 1.22      & 1.06      & 1.12      & 1.00      & 1.03      & 1.13 & \underline{\textbf{1.04}} & 1.33 & 1.03 & 1.02      & 0.75 \\
CSF (R) & 0.95      & 1.32      & 1.08      & 1.06      & 1.09      & 0.98      & 1.22 & 1.05 & 1.29 & 1.17 & 0.85      & 0.76 \\
Sym-RSfM   & \underline{\textbf{0.36}} & \underline{\textbf{0.43}} & \underline{\textbf{0.30}} & \underline{\textbf{0.43}} & \underline{\textbf{0.31}} & \underline{\textbf{0.26}} & \underline{\textbf{0.32}} & \underline{\textbf{1.04}} & \underline{\textbf{0.25}} & \underline{\textbf{0.16}} & \underline{\textbf{0.67}} & \underline{\textbf{0.26}} \\
\hline
\end{tabular*}

\begin{tabular*}{\textwidth}{@{\extracolsep{\fill}} || l || c | c | c | c || c  | c | c | c || c | c | c | c ||}
\hline
& \multicolumn{4}{c||}{\textbf{sofa}} & \multicolumn{4}{c||}{\textbf{train}} & \multicolumn{4}{c||}{\textbf{tv}} \\
\hline
& III & IV & V & VI  & I & II & III & IV  & I & II & III & IV   \\
\hline
RSfM  & 1.90 & 1.00 & 1.99 & 1.90 & 1.95 & 1.44 & 1.33 & 1.01 &  0.86 & 0.38 & 0.39 & 1.38 \\
CSF (S) & 1.16      & 0.99      & 1.66      & 1.21      & 0.84      & 0.69      & 0.86 & 0.85 & 0.99 & 0.79 & 0.95 & 0.73 \\
CSF (R) & 0.87      & 0.86      & \underline{\textbf{0.98}}      & 1.70      & 0.92      & \underline{\textbf{0.67}}      & \underline{\textbf{0.84}} & \underline{\textbf{0.82}} & 1.00 & 0.82 & 0.91 & 0.84 \\
Sym-RSfM  & \underline{\textbf{0.11}} & \underline{\textbf{0.69}} & 1.57 & \underline{\textbf{0.97}} & \underline{\textbf{0.18}} & 0.68 & 0.88 & 0.97 & \underline{\textbf{0.23}} & \underline{\textbf{0.14}} & \underline{\textbf{0.26}} & \underline{\textbf{0.44}} \\
\hline
\end{tabular*}
\caption{The mean \emph{rotation} errors for \emph{aeroplane, bus, car, sofa, train, tv}, calculated using the images from the same subtype (denoted by the Roman numerals) as input. 
}
\label{table:Rot_Results}
\end{table*}

\begin{table*}[t!]
\fontsize{9pt}{0.9\baselineskip}\selectfont
\begin{tabular*}{\textwidth}{@{\extracolsep{\fill}} || l || c | c | c | c | c | c | c || c | c | c | c | c | c ||}
\hline
& \multicolumn{7}{c||}{\textbf{aeroplane}} & \multicolumn{6}{c||}{\textbf{bus}} \\
\hline
& I & II & III & IV & V & VI & VII &  I & II & III & IV & V & VI  \\
\hline
RSfM & 0.44 & 1.17 & 0.52 & \underline{\textbf{0.29}} & 0.76 & 0.54 & 0.61 & 1.29 & 1.27 & 1.16 & 0.97 & 1.52 & 1.21 \\
CSF (S) & 1.25 & \underline{\textbf{0.36}} & 1.42 & 0.84 & 0.33 & 0.47 & \underline{\textbf{0.59}} & 1.11 & \underline{\textbf{0.39}} & 0.56 & \underline{\textbf{0.16}} & 3.02 & \underline{\textbf{0.44}} \\
CSF (R) & 0.25 & 0.44 & 0.34 & 1.40 & 0.58 & 1.73 & 0.69 & 0.99 & 1.06 & 1.33 & 0.88 & 1.94 & 2.00 \\
Sym-RSfM      & \underline{\textbf{0.19}} & 0.88 & \underline{\textbf{0.27}} & 0.34 & \underline{\textbf{0.33}} & \underline{\textbf{0.30}} & 0.62 & \underline{\textbf{0.68}} & 0.58 & \underline{\textbf{0.35}} & 0.24 & \underline{\textbf{0.76}} & 0.47 \\
\hline
\end{tabular*}

\begin{tabular*}{\textwidth}{@{\extracolsep{\fill}} || l || c | c | c | c | c | c | c | c | c | c || c | c ||}
\hline
& \multicolumn{10}{c||}{\textbf{car}} & \multicolumn{2}{c||}{\textbf{sofa}} \\
\hline
& I & II & III & IV & V & VI & VII & VIII & IX & X  & I & II  \\
\hline
RSfM & 1.48 & 1.49 & 1.33 & 1.38 & 1.45 & 1.39 & 1.21 & 1.81 & 1.22 & 1.07 & 2.50 & 1.09  \\
CSF (S) & 1.06 & 2.33 & 1.15 & 1.17 & 1.36 & 1.17 & 1.03 & 1.10 & 2.03 & 0.99 & 1.78 & 0.24 \\
CSF (R) & 1.34 & 1.07 & 1.03 & 1.16 & 1.18 & 1.26 & 0.88 & \underline{\textbf{0.90}} & 1.65 & 1.13 & \underline{\textbf{0.76}} & 0.25 \\
Sym-RSfM    & \underline{\textbf{1.03}} & \underline{\textbf{0.96}} & \underline{\textbf{0.95}} & \underline{\textbf{1.07}} & \underline{\textbf{0.89}} & \underline{\textbf{1.00}} & \underline{\textbf{0.81}} & 1.66 & \underline{\textbf{0.88}} & \underline{\textbf{0.71}} & 2.27 & \underline{\textbf{0.22}} \\
\hline
\end{tabular*}

\begin{tabular*}{\textwidth}{@{\extracolsep{\fill}} || l || c | c | c | c || c  | c | c | c || c | c | c | c ||}
\hline
& \multicolumn{4}{c||}{\textbf{sofa}} & \multicolumn{4}{c||}{\textbf{train}} & \multicolumn{4}{c||}{\textbf{tv}} \\
\hline
& III & IV & V & VI  & I & II & III & IV  & I & II & III & IV   \\
\hline
RSfM & 1.49 & 1.60 & 3.44 & 2.56 & 1.68 & 0.39 & 0.28 & 0.22 & 0.23 & 0.88 & 0.64 & 1.77 \\
CSF (S) & 3.14 & 1.54 & 2.74 & 1.55 & 0.83 & 0.85 & 0.25 & 0.26 & 0.66 & 0.77 & 0.34 & 0.34 \\
CSF (R) & 1.82 & 1.19 & 1.42 & 1.20 & 1.05 & \underline{\textbf{0.37}} & 0.24 & \underline{\textbf{0.17}} & 0.22 & 0.97 & 0.55 & 0.36 \\
Sym-RSfM  & \underline{\textbf{0.40}} & \underline{\textbf{1.07}} & \underline{\textbf{0.87}} & \underline{\textbf{1.14}}  & \underline{\textbf{0.73}} & 0.61 & \underline{\textbf{0.13}} & 0.24 & \underline{\textbf{0.09}} & \underline{\textbf{0.29}} & \underline{\textbf{0.32}} & \underline{\textbf{0.14}} \\
\hline
\end{tabular*}
\caption{The mean \emph{shape} errors for \emph{aeroplane, bus, car, sofa, train, tv}, calculated using the images from the same subtype (denoted by the Roman numerals) as input.}
\label{table:Shp_Results}
\end{table*}

\vspace{-2mm}
\subsection{Experiments on The Symmetric Rigid Structure from Motion}  \label{Rigid_Exp}
\vspace{-1mm}
We estimate the 3D structures of each subtype and the orientations of all the images within that subtype for \emph{aeroplane, bus, car, sofa, train, tv} in Pascal3D+ \cite{Xiang14}. Note that Pascal3D+ provides a single 3D shape for each subtype rather than for each object. For example, it provides 10 subtypes for the \emph{car} category, such as \emph{sedan, truck}, but ignores the within-subtype variation \cite{gao2016semi}. Thus, we divide the images of the same category into subtypes, and then input the images of each subtype for the experiments.

Following \cite{Kar15,Gao16}, images with more than 5 visible keypoints are used. The rotation and shape errors are calculated by Eq. \eqref{eq:Error}. The rigid SfM (RSfM) \cite{Tomasi92} and a more recent CSF method \cite{Gotardo11}, which both do not exploit symmetry, are used for comparison. Note that the CSF method \cite{Gotardo11} utilized smooth time-trajectories as initialization, which does not always
hold in our application, as the input images here are not from a continuous video. Thus, we also investigate the results from CSF method with random initialization. We report the CSF results with smooth prior as \emph{CSF (S)} and the \emph{best} results with 10 random initialization as \emph{CSF (R)}.

The results (mean rotation and shape errors) are shown in Tables \ref{table:Rot_Results} and \ref{table:Shp_Results}, which indicate that our method outperforms the baseline methods for most cases. The cases that our method does not perform as the best may be caused by that Pascal3D+ assumes the shapes from objects within the same subtype are very similar to each other, but this might be violated sometimes. Moreover, our method is robust to imperfect annotations (\ie, result in imperfect symmetric pairs) for practical use. This was simulated by adding Gaussian noise to the 2D annotations in the supplementary material.

\vspace{-2mm}
\section{Conclusions}
\vspace{-2mm}
We show that symmetry, Manhattan and multiple-image cues can be utilized to achieve good quality performance on object 3D structure reconstruction. For the single image case, symmetry and Manhattan together are sufficient if we can identify suitable keypoints. For the multiple-image case, we formulate the problem in terms of energy minimization exploiting symmetry, and optimize it by a coordinate descent algorithm. To initialize this algorithm, we define a surrogate energy function exploiting symmetry, and decompose it into a sum of two independent terms that can be solved by SVD separately. We further study the ambiguities of the surrogate energy and show that they can be resolved assuming the orthographic projection. Our results outperform the baselines on most object classes in Pascal3D+. Future works involve using richer camera models, like perspective \cite{park20153d,hartley2008perspective}, and keypoints extracted and matched automatically from images \cite{chen2014articulated,ma2015robust,ma2013regularized} with outliers \cite{ma2015non,ma2014robust,ma2015robust2} and occlusions \cite{jacobs2001linear,jacobs1997linear,Lin2010} handled.
~\\
~\\
\noindent \textbf{Acknowledgments.} We would like to thank Ehsan Jahangiri, Wei Liu, Cihang Xie, Weichao Qiu, Xuan Dong, and Siyuan Qiao for giving feedbacks on this paper. This work is partially supported by ONR N00014-15-1-2356 and the NSF award CCF-1317376.

\bibliographystyle{ieee}
\bibliography{egbib2}



\section*{Supplementary material}

# Supplementary Material for the Paper:

# Exploiting Symmetry and/or Manhattan Properties for 3D Object Structure Estimation from Single and Multiple Images

Yuan Gao
Tencent AI Lab, Shenzhen, China
ethanygao@tencent.com

Alan L. Yuille
Johns Hopkins University, Baltimore, MD
UCLA, Los Angeles, CA
alan.yuille@jhu.edu

This supplementary material contains more details of the Paper "Exploiting Symmetry and/or Manhattan Properties for 3D Object Structure Estimation from Single and Multiple Images":

1. In Section S1, we show our results on the imperfect annotations (*i.e.* the imperfect symmetric pairs).

2. In Section S2, we discuss how to impose Manhattan constraints as a regularization term for the Symmetric Rigid Structure from Motion (Sym-RSfM) on multiple images.

3. In Section S3, the minimization of the energy function w.r.t. the camera projection matrix $R_n$ under orthogonality constraints is detailed.

4. In Section S4, we provide a way to recover a $2 \times 2$ matrix $B$ from $BB^T$ up to a rotation ambiguity.

## S1. Experimental Results on The Imperfect Annotations

In this section, we investigate what happens if the keypoints are not perfectly annotated. This is important to check because our method depends on keypoint pairs therefore may be sensitive to errors in keypoint location, which will inevitably arise when we use features detectors, *e.g.* deep nets [1], to detect the keypoints.

To simulate this, we add Gaussian noise $\mathcal{N}(0, \sigma^2)$ to the 2D annotations and re-do the experiments. The standard deviation is set to $\sigma = sd_{max}$, where $d_{max}$ is the longest distance between all the keypoints (*e.g.* for *car*, it is the distance between the left/right front wheel to the right/left back roof top). We have tested for different $s$ by: 0.03, 0.05, 0.07. Other experimental settings are the same as them in the main text, *i.e.* images with more than 5 visible keypoints are used.

The mean rotation errors and the mean shape errors for *car* with $s = 0.03, 0.05, 0.07$ are shown in Tables S1 and S2. Each result value is obtained by averaging 10 repetitions. The results in Tables S1 and S2 show that the performances of all the methods decrease in general with the increase in the noise level. Nonetheless, our methods still outperform our counterparts with the noisy annotations (*i.e.* the imperfectly labeled annotations).

In summary, this section certificates our method is robust to imperfect annotations for practical use.

## S2. Imposing Manhattan Constraints to The Symmetric Rigid Structure from Motion

The energy function *w.r.t* $R_n, S$ (when the missing points are fixed) is:

$$\begin{aligned}
\mathcal{Q}(R_n, S) &= -\sum_n \ln P(Y_n, Y_n^\dagger | R_n, S) = -\sum_n \left( \ln P(Y_n | R_n, S) - \ln P(Y_n^\dagger | R_n, S) \right) \\
&= \sum_n ||Y_n - R_n S||_2^2 + \sum_n ||Y_n^\dagger - R_n \mathcal{A} S||_2^2 + Constant \\
&= \sum_n ||\mathbb{Y} - G_n \mathbb{S}||_2^2 + \sum_n ||\mathbb{Y}^\dagger - G_n \mathcal{A}_P \mathbb{S}||_2^2 + Constant
\end{aligned} \quad \text{(S1)}$$

1

|        | $\sigma = 0.03\, d_{max}$ |     |     |     |     |     |     |     |     |     | $\sigma = 0.05\, d_{max}$ |     |     |     |     |
|--------|------|------|------|------|------|------|------|------|------|------|------|------|------|------|------|
|        | I    | II   | III  | IV   | V    | VI   | VII  | VIII | IX   | X    | I    | II   | III  | IV   | V    |
| RSfM   | 0.57 | 0.69 | 0.55 | 1.09 | 0.68 | 1.58 | 0.67 | 1.45 | 0.97 | 0.37 | 0.59 | 0.74 | 0.53 | 1.11 | 0.68 |
| CSF (S)| 0.95 | 1.22 | 1.10 | 1.01 | 1.01 | 1.03 | 1.16 | **1.02** | 1.35 | 1.04 | 0.95 | 1.29 | 1.05 | 1.02 | 1.00 |
| CSF (R)| 0.95 | 1.27 | 1.07 | 1.05 | 1.05 | 0.98 | 0.95 | 1.04 | 1.05 | 1.16 | 0.95 | 1.24 | 1.07 | 1.04 | 1.07 |
| Sym-RSfM | **0.37** | **0.42** | **0.34** | **0.46** | **0.32** | **0.26** | **0.32** | 1.06 | **0.25** | **0.12** | **0.39** | **0.45** | **0.31** | **0.48** | **0.31** |

|        | $\sigma = 0.05\, d_{max}$ (cont.) |     |     |     |     | $\sigma = 0.05\, d_{max}$ |     |     |     |     |     |     |     |     |     |
|--------|------|------|------|------|------|------|------|------|------|------|------|------|------|------|------|
|        | VI   | VII  | VIII | IX   | X    | I    | II   | III  | IV   | V    | VI   | VII  | VIII | IX   | X    |
| RSfM   | 1.54 | 0.69 | 1.49 | 0.99 | 0.40 | 0.59 | 0.72 | 0.56 | 1.09 | 0.64 | 1.54 | 0.67 | 1.55 | 0.97 | 0.38 |
| CSF (S)| 1.04 | 1.09 | **1.03** | 1.37 | 1.02 | 1.04 | 1.25 | 1.05 | 1.16 | 0.98 | 1.03 | 0.88 | 1.02 | 1.20 | 1.04 |
| CSF (R)| 0.97 | 0.93 | **1.03** | 0.89 | 1.16 | 0.95 | 1.22 | 1.06 | 1.04 | 0.88 | 0.96 | 1.00 | 1.04 | 0.91 | 1.05 |
| Sym-RSfM | **0.28** | **0.33** | 1.13 | **0.28** | **0.13** | **0.38** | **0.37** | **0.30** | **0.49** | **0.28** | **0.35** | **0.33** | 1.01 | **0.26** | **0.13** |

Table S1. The mean *rotation* errors for *car* with imperfect annotations. The noise is Gaussian $\mathcal{N}(0, \sigma^2)$ with $\sigma = s d_{max}$, where we choose $s = 0.03, 0.05, 0.07$ and $d_{max}$ is the longest distance between all the keypoints (*i.e.* the left/right front wheel to the right/left back roof top). The Roman numerals denotes the index of the subtype. Each result value is obtained by averaging 10 repetitions.

|        | $\sigma = 0.03\, d_{max}$ |     |     |     |     |     |     |     |     |     | $\sigma = 0.05\, d_{max}$ |     |     |     |     |
|--------|------|------|------|------|------|------|------|------|------|------|------|------|------|------|------|
|        | I    | II   | III  | IV   | V    | VI   | VII  | VIII | IX   | X    | I    | II   | III  | IV   | V    |
| RSfM   | 1.48 | 1.48 | 1.33 | 1.37 | 1.45 | 1.39 | 1.21 | 1.82 | 1.23 | 1.07 | 1.48 | 1.47 | 1.34 | 1.39 | 1.44 |
| CSF (S)| 1.06 | 2.36 | 1.14 | **0.88** | 1.37 | 1.17 | **0.77** | 1.13 | 2.00 | 0.98 | 1.10 | 1.22 | 1.15 | **0.90** | 1.34 |
| CSF (R)| 1.33 | 0.99 | 1.02 | 1.15 | 1.18 | 1.25 | 0.87 | **0.90** | 1.42 | 1.12 | 1.33 | 0.98 | 1.01 | 1.15 | 1.16 |
| Sym-RSfM | **1.04** | **0.95** | **0.96** | 1.08 | **0.90** | **1.12** | 0.81 | 1.80 | **0.88** | **0.66** | **1.04** | **0.95** | **0.95** | 1.08 | **0.90** |

|        | $\sigma = 0.05\, d_{max}$ (cont.) |     |     |     |     | $\sigma = 0.05\, d_{max}$ |     |     |     |     |     |     |     |     |     |
|--------|------|------|------|------|------|------|------|------|------|------|------|------|------|------|------|
|        | VI   | VII  | VIII | IX   | X    | I    | II   | III  | IV   | V    | VI   | VII  | VIII | IX   | X    |
| RSfM   | 1.43 | 1.20 | 1.81 | 1.21 | 1.08 | 1.48 | 1.46 | 1.33 | 1.36 | 1.43 | 1.48 | 1.21 | 1.79 | 1.22 | 1.08 |
| CSF (S)| **1.20** | 1.04 | 1.11 | 1.96 | 0.97 | 3.56 | 1.28 | 1.15 | 1.19 | 1.38 | 1.22 | 2.08 | 1.06 | 2.47 | 1.03 |
| CSF (R)| 1.25 | 0.87 | **0.88** | 1.41 | 1.11 | 1.31 | **0.92** | 0.99 | 1.15 | 1.17 | 1.25 | 0.87 | **0.88** | 1.55 | 1.11 |
| Sym-RSfM | 1.25 | **0.81** | 1.71 | **0.88** | **0.67** | **1.05** | 0.95 | **0.96** | **1.09** | **0.89** | **1.07** | **0.82** | 1.80 | **0.88** | **0.68** |

Table S2. The mean *shape* errors for *car* with imperfect annotations. Other parameters are the same as Table S1.

where $\mathbb{S} \in \mathbb{R}^{3P\times 1}, \mathbb{Y}_n \in \mathbb{R}^{2P\times 1}, \mathbb{Y}_n^\dagger \in \mathbb{R}^{2P\times 1}$ are vectorized $S, Y_n, Y_n^\dagger$, respectively. $\mathcal{A} = \text{diag}[-1, 1, 1]$ is a matrix operator which negates the first row of its right-multiplied matrix. $G_n = I_P \otimes R_n$ and $\mathcal{A}_P = I_P \otimes \mathcal{A}$. $I_P \in \mathbb{R}^{P \times P}$ is an identity matrix. The last equation is obtained by vectorizing the above one by $\text{vec}(AXB^T) = (B \otimes A)\text{vec}(X)$, and $\otimes$ denotes Kronecker product.

If another Manhattan direction is available *e.g.* if we have $S_i = [S_i^x, S_i^{y_i}, S_i^z]^T, S_j = [S_i^x, S_i^{y_j}, S_i^z]^T$ along $y$-axis, *i.e.* $S_i, S_j$ have the same coordinates on $x$- and $z$-axes[1]. Then, we have another constraints: $S_i - S_j = [0, S_i^{y_i} - S_i^{y_j}, 0]^T$. Let the matrix operator $\mathcal{X} = \text{diag}([1,0,1])$ which selects the first and the third row of its right-multiplied matrix, we can rewrite Eq. (S1) by encoding the Manhattan constraints as regularization:

$$\mathcal{Q}(\mathbb{S}) = \sum_n^N \left( ||\mathbb{Y} - G_n \mathbb{S}||_2^2 + ||\mathbb{Y}^\dagger - G_n \mathcal{A}_P \mathbb{S}||_2^2 \right) + \lambda ||\mathcal{X}(S_i - S_j)||_2^2. \tag{S2}$$

We can further rewrite the last term of Eq. (S2) based on $\mathbb{S}$ (instead $S_i, S_j$) by applying matrix operators on $S$ and then vectorizing it. We define the matrix operator $\mathcal{Y} \in \mathbb{R}^{P\times 2}$ such that its $i$'th row in first column and $j$'th row in second column are equal to 1, and otherwise 0. Then, $S\mathcal{Y}$ selects the keypoints $S_i, S_j$ ($i$'th and $j$'th columns of $S$) along the Manhattan direction $y$. We also use the matrix operator $\mathcal{Z} = [-1, 1]^T \in \mathbb{R}^{2\times 1}$ so that $S\mathcal{Y}\mathcal{Z}$ denotes $S_j - S_i$. Thus, Eq. (S2) can be written by:

$$\mathcal{Q}(S) = \sum_n^N \left( ||\mathbb{Y} - G_n \mathbb{S}||_2^2 + ||\mathbb{Y}^\dagger - G_n \mathcal{A}_P \mathbb{S}||_2^2 \right) + \lambda ||\mathcal{X}S\mathcal{Y}\mathcal{Z}||_2^2.$$

$$= \sum_n^N \left( ||\mathbb{Y} - G_n \mathbb{S}||_2^2 + ||\mathbb{Y}^\dagger - G_n \mathcal{A}_P \mathbb{S}||_2^2 \right) + \lambda ||(\mathcal{Z}^T \mathcal{Y}^T \otimes \mathcal{X})\mathbb{S}||_2^2. \tag{S3}$$

---
[1] $S_i$ and $S_j$ are not necessary to be symmetric along $y$-axis.

Taking derivative the energy function in Eq. (S2) *w.r.t* $\mathbb{S}$ and equating it to 0, the update of $\mathbb{S}$ under additional Manhattan constraint becomes:

$$\mathbb{S} = \left(\sum_{n=1}^{N}(G_n^T G_n + \mathcal{A}_P^T G_n^T G_n \mathcal{A}_P^T) + \lambda(\mathcal{Z}^T \mathcal{Y}^T \otimes \mathcal{X})^T(\mathcal{Z}^T \mathcal{Y}^T \otimes \mathcal{X})\right)^{-1} \left(\sum_{n=1}^{N}(G_n^T \mathbb{Y}_n + \mathcal{A}_P^T G_n^T \mathbb{Y}_n^\dagger)\right). \quad \text{(S4)}$$

Note that it is easy to generalize the matrix operator $\mathcal{Y}, \mathcal{Z}$ if we have $M$ points along $y$-axis. Specifically, we only need to rewrite $\mathcal{Y} \in \mathbb{R}^{P \times M}$ that selects the $M$ keypoints ($M$ columns of $S$) along the Manhattan direction $y$, and $\mathcal{Z} = [-\mathbf{1}_{M-1}, I_{M-1}]^T \in \mathbb{R}^{M \times M-1}$ that makes each of the $M$ keypoints minus the first keypoint.

## S3. Update Camera Projection Matrix $R_n$ Under Orthogonality Constraints

In this section, we describe how to do coordinate descent for the camera parameters $R_n$ given the 3D structure $S$ is fixed.

In order to minimize Eq. (S1) *w.r.t.* $R_n$ under the nonlinear orthogonality constraints $R_n R_n^T = I$, we follow an alternative approach used in [2] to parameterize $R_n$ as a complete $3 \times 3$ rotation matrix $Q_n$ and update the incremental rotation on $Q_n$ instead, *i.e.* $Q_n^{new} = e^\xi Q_n$.

Here, the first and second rows of $Q_n$ is the same as $R_n$, and the third row of $Q_n$ is obtained by the cross product of its first and second rows. The relationship of $Q_n$ and $R_n$ can be revealed by a matrix operator $\mathcal{M}$:

$$R_n = \mathcal{M} Q_n, \qquad \mathcal{M} = \begin{bmatrix} 1, 0, 0 \\ 0, 1, 0 \end{bmatrix}. \quad \text{(S5)}$$

Note that the incremental rotation $e^\xi$ can be further approximated by its first order Taylor Series, *i.e.* $e^\xi \approx I + \xi$. Finally, we have:

$$R_n^{new}(\xi) = \mathcal{M}(I + \xi) Q_n. \quad \text{(S6)}$$

Therefore, setting $\partial \mathcal{Q}/\partial R_n = 0$, then replace $R_n$ by $Q_n$ using Eq. (S6) and vectorize it, we have:

$$R_n = \mathcal{M} e^\xi Q_n \approx \mathcal{M}(I + \xi) Q_n \qquad \text{and} \qquad \text{vec}(\xi) = \alpha^+ \beta,$$

$$\alpha = \left(\sum_{p=1}^{P}(S_p S_p^T + \mathcal{A} S_p S_p^T \mathcal{A}^T)^T Q_n^T\right) \otimes \mathcal{M},$$

$$\beta = \text{vec}\left(\sum_{p=1}^{P}(Y_{n,p} S_p^T + Y_{n,p}^\dagger S_p^T \mathcal{A}^T) - Q_n \sum_{p=1}^{P}(S_p S_p^T + \mathcal{A} S_p S_p^T \mathcal{A}^T)\right), \quad \text{(S7)}$$

where the subscript $p$ means the $p$th keypoint, $\alpha^+$ means the pseudo inverse matrix of $\alpha$, $\otimes$ denotes Kronecker product.

## S4. Recover Matrix $B$ from $BB^T$

In the following, we describe a method for recovering $2 \times 2$ matrix $B$ from $BB^T$ up to a 2D rotation. Let $B = \begin{bmatrix} b_1 \cos\theta, & b_1 \sin\theta \\ b_3 \cos\phi, & b_3 \sin\phi \end{bmatrix}$, thus:

$$BB^T = \begin{bmatrix} (b_1)^2, & b_1 b_3 \cos(\theta - \phi) \\ b_1 b_3 \cos(\theta - \phi), & (b_3)^2 \end{bmatrix} = \begin{bmatrix} bb_1, & bb_2 \\ bb_2, & bb_3 \end{bmatrix}. \quad \text{(S8)}$$

If we assume $\phi = 0$ (due to the "fake" rotation ambiguity on $yz$-plane) and $b_1, b_3 \geq 0$ (due to the "fake" direction ambiguities of $y-$ and $z-$ axes), all the unknown parameters (*i.e.* $b_1, b_3, \theta$) can be calculated by:

$$b_1 = \sqrt{bb_1}, \qquad b_3 = \sqrt{bb_3}, \qquad \theta = \text{arc}\cos(\frac{bb_2}{b_1 b_3}) + \phi, \qquad \phi = 0. \quad \text{(S9)}$$



\end{document}